\documentclass[runningheads]{llncs}

\usepackage{graphicx}
\usepackage{comment}
\usepackage{amsmath,amssymb}
\usepackage{color}

\usepackage[american]{babel}
\usepackage{hyperref}

\addto\extrasamerican{
}

\newif\ifreview
\reviewtrue
\reviewfalse

\ifreview
	\usepackage{lineno}

	\linenumbers
\fi

\usepackage{xspace}
\newcommand{\etal}{\textit{et al.}\xspace}
\newcommand{\repeatthanks}{\textsuperscript{\thefootnote}}

\begin{document}


\def\SubNumber{000}

\def\GCPRTrack{Regular Track}

\title{Spatial Transformer Networks for Curriculum Learning}

\ifreview
	\titlerunning{DAGM GCPR 2021 Submission \SubNumber{}. CONFIDENTIAL REVIEW COPY.}
	\authorrunning{DAGM GCPR 2021 Submission \SubNumber{}. CONFIDENTIAL REVIEW COPY.}
	\author{DAGM GCPR 2021 - \GCPRTrack{}}
	\institute{Paper ID \SubNumber}
\else

\author{
Fatemeh Azimi\thanks{equal contribution}\textsuperscript{1,2} \and 
Jean-Francois Jacques Nicolas Nies\repeatthanks\textsuperscript{1,2} \and 
Sebastian Palacio\textsuperscript{1} \and
Federico Raue\textsuperscript{1} \and J\"orn Hees\textsuperscript{1} \and Andreas Dengel\textsuperscript{1,2}
}
\institute{\textsuperscript{1} TU Kaiserslautern, Germany\\
}

	
	\authorrunning{F. Azimi et al.}

\institute{\textsuperscript{1} TU Kaiserslautern, Germany\\
\textsuperscript{2}Smart Data and Knowledge Services, German Research Center for Artificial Intelligence (DFKI), Germany\\
\texttt{\{fatemeh.azimi, jean-francois\_jacques\_nicolas.nies, sebastian.palacio, federico.raue, joern.hees, andreas.dengel\}@dfki.de}}
\fi

\maketitle              

\begin{abstract}
Curriculum learning is a bio-inspired training technique that is widely adopted to machine learning for improved optimization and better training of neural networks regarding the convergence rate or obtained accuracy.
The main concept in curriculum learning is to start the training with simpler tasks and gradually increase the level of difficulty. 
Therefore, a natural question is how to determine or generate these simpler tasks.
In this work, we take inspiration from Spatial Transformer Networks (STNs) in order to form an easy-to-hard curriculum. 
As STNs have been proven to be capable of removing the clutter from the input images and obtaining higher accuracy in image classification tasks, we hypothesize that images processed by STNs can be seen as easier tasks and utilized in the interest of curriculum learning.
To this end, we study multiple strategies developed for shaping the training curriculum, using the data generated by STNs.
We perform various experiments on cluttered MNIST and Fashion-MNIST datasets, where on the former, we obtain an improvement of $3.8$pp in classification accuracy compared to the baseline.

\end{abstract}
\section{Introduction}
The learning process of intelligent beings such as humans and animals depends on many factors, one of which is the method of teaching and the way the information is presented to the learner.
Numerous findings in the area of cognitive science demonstrate the importance of teaching designs and shaping the way the tasks are presented to the student~\cite{skinner1958reinforcement,peterson2004day}.

Inspired by these findings on the learning mechanisms of humans, researchers in the machine learning community have explored the potential of utilizing similar ideas for better training the neural networks~\cite{elman1993learning,bengio2009curriculum,sanger1994neural,selfridge1985training}. Better training can be either concerning faster convergence, finding better local minima, improved generalization, etc.
This line of research is referred to as curriculum learning.

There is a wide range of approaches addressing various aspects of this field in different applications.
Rohde \etal~investigated the effect of starting from easier examples in the task of language acquisition~\cite{rohde1999language} and in~\cite{bengio2009curriculum}, the authors studied where and when curriculum learning can be beneficial for the training process in the context of shape recognition tasks.
Erhan \etal~explored the impact of unsupervised pretraining as a way of curriculum learning~\cite{erhan2009difficulty}. 
They showed that the pretraining step allows the optimization process to reach a local minimum where the final test error is significantly better compared to directly training on the supervised task.
Numerous efforts have looked into the impact of hard mining~\cite{chang2017active,zhang2017spftn,shrivastava2016training,schein2007active}, which is finding the more challenging samples from the network's perspective and assigning them a higher weight.
In~\cite{weinshall2018curriculum,hacohen2019power}, the authors proposed multiple ways to sort the data samples based on an increase in difficulty.
The goal is to measure the hardness of each task and develop a schedule or sampling method that initially exposes easier samples to the network.
In the Teacher-Student training strategy~\cite{hinton2015distilling}, two agents are trained together in a way that the base learner (student network) is responsible for learning the actual task while the other agent (teacher network) learns to optimize certain aspects of the training process such as data sampling for the student~\cite{matiisen2019teacher}. 
In this regard, \cite{fan2018learning} studies the question of what constitutes the characteristics of a good teacher in the curriculum learning framework.

Unlike the previous methods that focus on developing methods for distinguishing between the easy and hard samples~\cite{hacohen2019power,weinshall2018curriculum}, we intend to directly manipulate the data distribution and generate the simpler tasks required for curriculum learning.
Motivated by the success of Spatial Transformer Networks (STNs)~\cite{jaderberg2015spatial,azimi2019reinforcement} in image classification, we take a new approach towards the preparation of easy and hard data samples for curriculum learning.
STNs learn affine transformations that modify input data such that it is \textit{easier} for the network to classify;
in this work, we develop an approach to utilize the learned transformations in favor of curriculum learning.
We highlight that by \textit{easier} data distribution, we mean one that obtains a higher classification accuracy when using an identical classifier architecture and network capacity.
We hypothesize that the transformation by STN simplifies all the intricacies of the original distribution that are caused by noise and clutter. The motivation is that the network first learns on data with a high signal-to-noise ratio in the hope that, when the hard, noisy samples come, the model will be inclined to ignore artifacts caused by the noise itself.
It is worth remarking that we utilize the data transformed by STN during the training, but do not use the STN during the test phase which allows us to have a simpler inference framework.

In this paper, we specifically use Sequential STNs (SSTNs)~\cite{azimi2019reinforcement}, as they allow us to access a range of data that varies in difficulty, from the original image to all incremental transformations thereof.
We introduce two main approaches that exploit SSTN-processed data. 
The first variant, called Mixed-batch training, includes a portion of the SSTN-processed data with each mini-batch during training.
The rationale is that the network can learn an embedding that is robust to the clutter of the original data by concurrently optimizing on the decluttered samples produced by the SSTN. 
In the second approach, called Incremental Difficulty training, optimization begins by exclusively using data that has been fully processed by the SSTN (i.e., data has undergone the maximum number of transformation steps).
Then, we gradually increase the difficulty of samples being used for training by decreasing the number of transformation steps that the SSTN applies to each image (zero steps corresponds to the original data).
We evaluate our proposed approach on cluttered MNIST and Fashion-MNIST datasets. The experimental results show that Spatial Transformers can indeed be used as a tool for forming the curriculum learning schedule.

\section{Related Work}
Several works in the field of cognitive science and animal training have demonstrated that humans and animals learn faster when the concepts are presented in an organized way with gradually increasing levels of difficulty \cite{peterson2004day,skinner1958reinforcement,pavlov2010conditioned}.
In~\cite{skinner1958reinforcement} the authors discuss the importance of successive reinforcement and examine the impact of various scheduling methods to accelerate the learning process in the case of birds.
Moreover, Kruger \etal~\cite{krueger2009flexible} study the effect of the shaping mechanism concerning the arrangement of the presented tasks based on the complexity factor. 
According to their analysis, training time increases proportionally with the task complexity; however, different ways of shaping and scheduling can considerably reduce this time.

Inspired by these findings,~\cite{bengio2009curriculum} proposes applying the same methodology to machine learning, which at best aims to imitate the learning process of humans.
They formulate curriculum learning as a training policy that favors easier samples at the beginning of the training and gradually increases complexity.
They illustrate the effectiveness of this approach in improving generalization, achieving faster convergence, and finding better local minima in language modeling and shape recognition tasks.
Zaremba \etal~\cite{zaremba2014learning} demonstrate the efficacy of curriculum learning in training recurrent neural networks for executing simple computer programs.
Their scheme is based on a gradual increase in the length and number of nested loops in the program.
To develop a curriculum learning based training strategy, \cite{hacohen2019power} proposes a sorting method to order the tasks based on an increase in hardness and in \cite{graves2017automated,graves2016hybrid} they utilize learning progress signals such as loss value or prediction gain to automatically select a training path for the neural network in order to enhance learning efficiency.
Similarly, in \cite{bengio2015scheduled,azimi2020hybrid}, the authors study the use of curriculum learning in sequence prediction.
They suggest replacing the ground-truth with the model's prediction following a probabilistic scheme.
When training is starting, the ground-truth labels are used for the model to predict the target in the next time steps;
then, gradually, the ground truth is replaced with the model's prediction, and the task's difficulty is increased.
In \cite{florensa2017reverse} the authors use curriculum learning in the context of goal-oriented Reinforcement Learning where an agent is expected to navigate an environment towards a target by reversing the task; instead of placing the agent at a random location and train it to find the goal state, they start from this state and progressively increase the distance, and hence, the task complexity.
In another work, Wang \etal use curriculum learning to address the challenges of training with imbalanced data where the dataset is not following an independent and identical distribution~\cite{wang2019dynamic}.
They propose a data sampling method that is responsible for sampling balanced samples and a loss scheduler that dynamically balances the weighting scheme between different loss components in their multi-task setup.

Another closely related line of work is the Teacher-Student training approach~\cite{hinton2015distilling} in which two agents are trained intertwined such that the teacher network conditions certain aspects of the student network's training such as the optimization objective or task complexity assortment~\cite{matiisen2019teacher}.
\cite{jiang2018mentornet} uses this training policy to mitigate challenges faced when training with noisy data and to prevent the network from overfitting on erroneous labels. 
To this end, they train a mentor network that generates a curriculum in the form of a weighting scheme for each training sample based on its importance and reliability. 
Subsequently, this weighting scheme is used for training the base classifier (student network) resulting in improved generalization.
In \cite{fan2018learning} the authors investigate what constitutes a good teacher in the context of training neural networks. 
They suggest a framework where teacher and student networks interact in a way where the student undergoes the standard training and the teacher receives a feedback signal from the student to decide about the training data, loss function, and the hypothesis space.

Different than the approaches discussed above, the focus of our proposed method is on generating the simpler tasks by deploying the Spatial Transformer Networks~\cite{jaderberg2015spatial,azimi2019reinforcement}.
We build on top of the well-developed recipe for curriculum learning (start simple!) and explore the potential of SSTNs for producing the easier tasks in the context of image classification.  

\section{Method}
In this section, we elaborate on the intuition behind deploying Sequential Spatial Transformer Networks for curriculum learning and proceed with explaining the proposed strategies concerning shaping the curriculum function. 

\subsection{Background}
\label{sec:bg}
Spatial Transformer Networks (STNs) were introduced in \cite{jaderberg2015spatial} as a method for improving the image classification accuracy.
The main idea in STNs is learning a function that generates the parameters of an affine transformation that modifies the input image such that it is easier to classify.
This function is estimated by a neural network and is trained by minimizing the classification loss.
According to the empirical results, the effect of applying the affine transformation in the context of image classification is focusing on the main content of the image which in turn makes the image \textit{easier} to classify.

\begin{figure}[h]
    \centering
    \includegraphics[width=\textwidth]{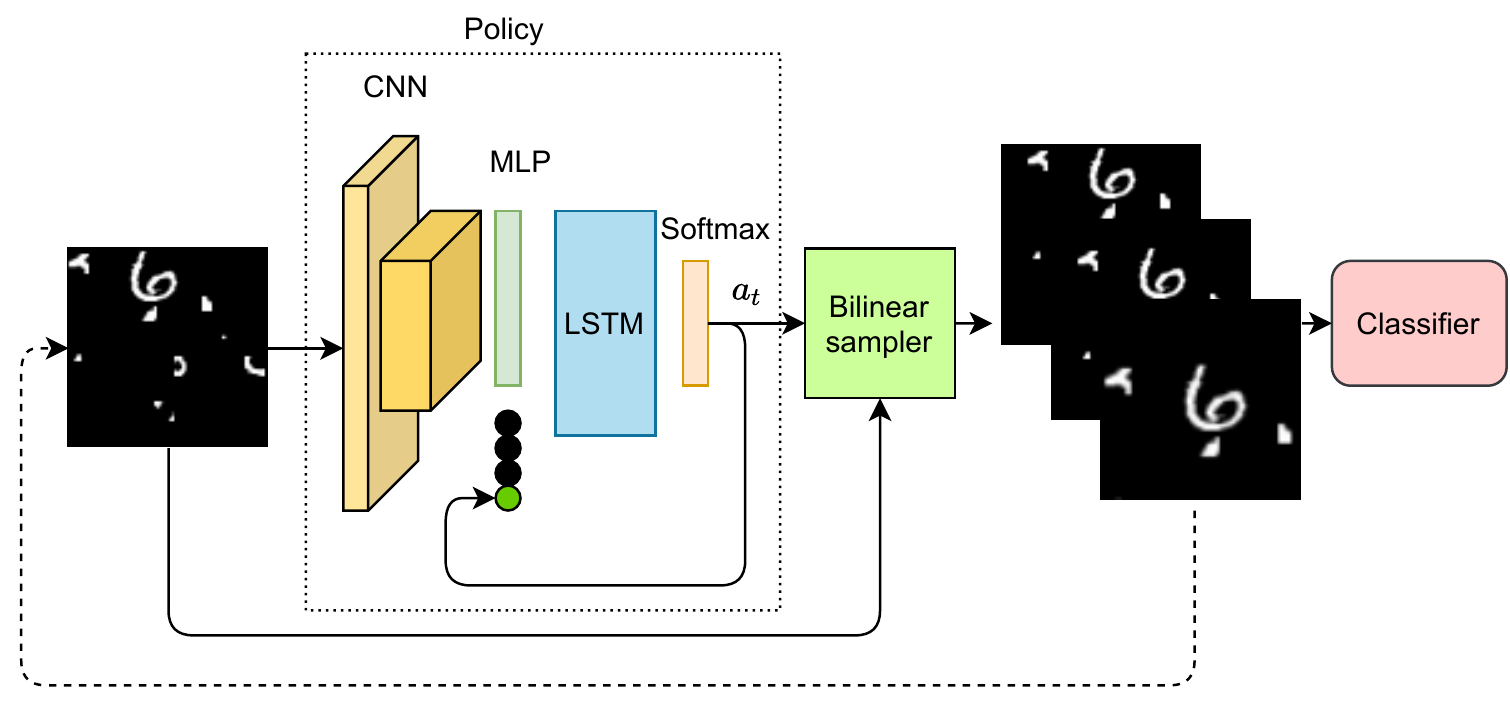}
    \caption{Architecture of the SSTN~\cite{azimi2019reinforcement}. At each time step the action $a_{t}$ (discrete transformation) is selected by the trained policy and the input content is gradually modified within a maximum of $T$ transformation steps.}
    \label{fig:SSTN}
\end{figure}

\begin{figure}[h]
    \centering
    \includegraphics[width=0.8\textwidth]{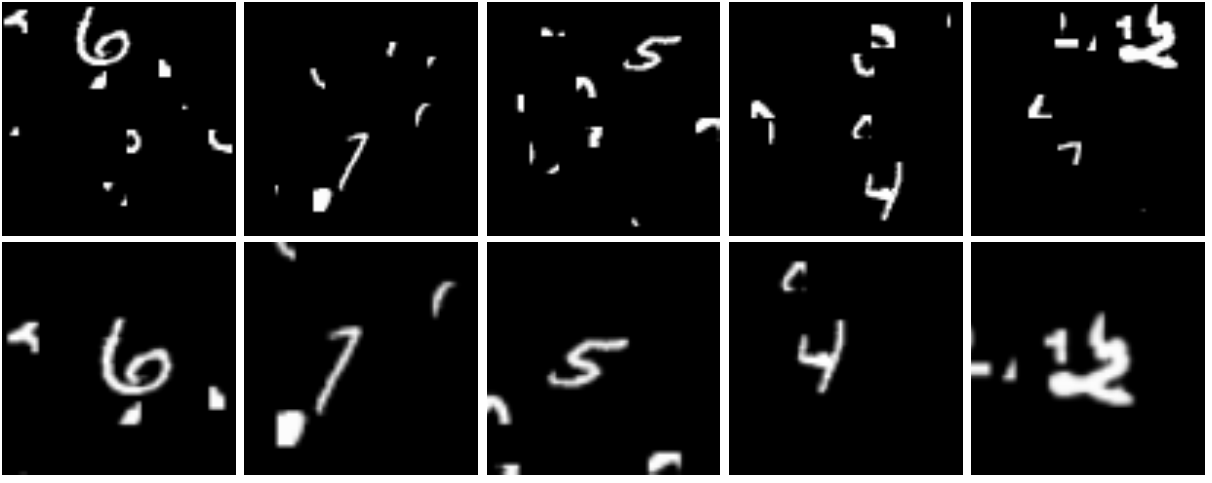}
    \caption{Data samples from cluttered MNIST dataset in the top row and the data transformed by SSTN in the second row.}
    \label{fig:before_afters}
\end{figure}

Following this idea, \cite{azimi2019reinforcement} propose the Sequential STN (SSTN) architecture.
In SSTN the authors suggest to instead of directly learning the parameters of the affine transformation, break the transformation into a sequence of simple and pre-defined transformations.
They consider a set of discrete transformations including vertical and horizontal translations, rotation, and scaling to constitute the final affine transformation.
Subsequently, they use Reinforcement Learning to train a policy that learns how to optimally combine these small transformations towards modifying the input image so that it is simpler to classify and results in a lower classification loss.
\autoref{fig:SSTN} illustrates the mechanism of the SSTN and \autoref{fig:before_afters} shows images before and after processing by the SSTN.

In this work, we build on top of the SSTN approach mainly due to its sequential nature.
Having a series of slightly modified and improved images at each time step, we hypothesize that this provides a hard-to-easy data distribution spectrum.
We aim to use this characteristic in a curriculum learning setup for improving the training process of the classifier.
In the following, we elaborate on two main approaches for employing the SSTN-processed data during the training, namely Mixed-batch and Incremental Difficulty training.

\subsection{Mixed-batch Curriculum Learning}
As discussed earlier, transforming the input data through SSTN shifts the data distribution of the input such that it is easier to classify (higher classification accuracy) compared to the original data.
Therefore, we measure the extent to which this easier distribution can be used as part of a curriculum for image classifiers.

In the mixed-batch strategy, we investigate if mixing the original input images with a portion of data transformed by SSTN can facilitate the training process of the classifier.
The hope is that by presenting the input samples from the SSTN-processed and original data, the classifier can learn a representation that ignores the existing clutter and picks on the main content of the image.
In this regard, we explore two different mixing strategies as described in the following.

In the first variant, we allocate a fixed portion of each mini-batch of input images to SSTN-processed data throughout the training.
Formally, assume $X = \{x_{i}\}_{i=1}^{N}$ refers to the original data distribution with $N$ data samples.
Consider $f_{t}$ as the transformation that is selected by the SSTN at the $t$th time-step and $x_{i}^{T}$ is the product of sequentially applying $T$ discrete transformations to $x_{i}$.
We refer to the modified data distribution as $\hat{X} = \{\hat{x_{i}}\}$ where $\hat{x_{i}} = f_{T} \circ f_{T-1} \ldots \circ f_{1} (x_{i})$ and $T$ is the maximum number of transformation steps.
In Mixed-batch setup, each mini-batch $B_{j}$ consists of a fixed portion from the original and the modified data distribution: $B_{j} \subset X \cup \hat{X}$.

Moreover, we experiment with the Mixed-batch idea in a slightly different fashion. 
Instead of using a fixed share of SSTN-processed data in each mini-batch, we start the training with each mini-batch fully transformed by SSTN; then, we gradually increase the ratio of original-to-easy samples as training progresses such that towards the end of the training, only the original data are used.
The rationale is to initially start with a high portion of \textit{easier} samples and incrementally move towards the original data distribution which will be used during the inference.
We refer to this scheme as Dynamic Mixed-batch training.

\subsection{Incremental Difficulty Curriculum Learning}
We note that the SSTN performs a series of small transformations towards the final processed results.
This allows us to have access to a spectrum of data distributions ranging from fully processed images (an easier distribution with higher classification accuracy) to original unprocessed data.

From the curriculum learning literature, we know that it is beneficial to the training of the neural networks to start from simpler tasks and progressively increase the complexity.
In Incremental difficulty setup, we identify a potential application of SSTN for generating this series of tasks or data distributions.
Having $T$ step transformations in SSTN, we start training the classifier with fully processed images and gradually decrease the number of transformation steps as the training progresses, towards the original data.
Formally, we sample each mini-batch $B_{j}$ from the dataset $X^{\prime}=\{f_{T} \circ f_{T-1} \ldots \circ f_{1} (x_{i})\}_{i=1}^{N}$ where the number of transformation steps $T$ gradually decreases from an initial maximum value to $0$, resulting in the original dataset $X$.   
In this regard, we experiment with various scheduling strategies including linear, cosine annealing, and exponential decay where each method changes the data distribution at a different rate.

\section{Experiments}
In this section, we present the experimental setup of our method, as well as the results from Mixed-batch, Dynamic Mixed-batch, and Incremental Difficulty curriculum training strategies.
First, we describe the datasets used in the experiments.
Then, we explain the experimental setup and analyze the obtained results in comparison with the baseline.
Finally, we provide an ablation study on the impact of difficulty scheduling functions and the number of applied transformations ($T$) in the SSTN policy.

\subsection{Datasets}
\begin{figure}[tbh]
    \centering
    \includegraphics[width=0.8\textwidth]{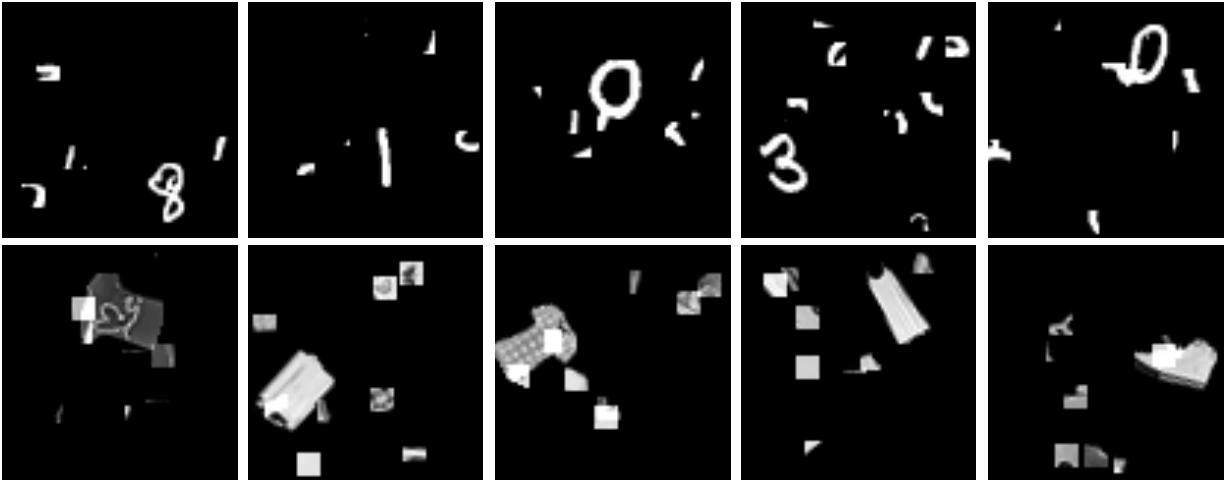}
    \caption{Data samples from the cluttered MNIST in the top row and Fahison-MNIST in the second row.}
    \label{fig:sample}
\end{figure}

Following the setup in~\cite{azimi2019reinforcement}, we evaluate our curriculum method on cluttered MNIST and Fashion-MNIST datasets.
Cluttered MNIST and cluttered Fashion-MNIST are two synthetic datasets generated from MNIST and Fashion MNIST datasets. 
Cluttered MNIST was introduced in~\cite{mnih2014recurrent} for the task of visual attention.
We use the public code\footnote{https://github.com/deepmind/mnist-cluttered} provided by~\cite{mnih2014recurrent} to synthesize these two datasets.
The images in the aforementioned datasets are grayscale and the image size is increased to $80\times80$.
The main content of the image (a number or a clothing item) undergoes an affine transformation and is placed at a random location within the image frame.
Furthermore, the image is cluttered with content-like appearances.
These modifications make the classification task significantly harder compared to the original MNIST and Fashion-Mnist datasets~\cite{azimi2019reinforcement}.
Visual samples from these datasets can be seen in \autoref{fig:sample}.

\subsection{Network Architecture, Setup and Experimental Results}
For our experimental setup, we follow the network architectures in \cite{azimi2019reinforcement}.
The policy network consists of three convolution layers followed by a LSTM~\cite{hochreiter1997long} module, a fully-connected layer, and softmax activation which generates the probability distribution of the optimal action selection.
The number of kernels in the convolution layers is $32$, $64$, and $64$ respectively.
The action-set employed in SSTN consists of $8$ distinct transformations, including a translation of $\pm4\ $ pixels in vertical and horizontal directions, scaling up with a factor of $1.2$, clockwise and counter-clockwise rotation of $10$ degrees, and Identity transformation.
Each sequence of transformations consists of $40$ transformation steps ($T=40$) unless mentioned otherwise, as we obtained the best results with this configuration.
We use the Reinforce algorithm to train the policy.
For more details on training the policy network, please refer to~\cite{azimi2019reinforcement}.

For the experiments on cluttered MNIST, the classifier architecture consists of a single convolution layer with $64$ kernels followed by two fully connected layers, referred to as LeNet1.
The classifier used for the experiments on cluttered Fashion-MNIST consists of two convolution layers with $32$ and $64$ kernels and two fully connected layers, referred to as LeNet2.
For all the experiments, we use the Adam optimizer~\cite{kingma2014adam}, a learning rate of $1e-4$, and a batch size of $64$.

\autoref{tab:main} presents the experimental results on cluttered MNIST and Fashion-MNIST datasets.
For the baseline, we trained the model with and without data augmentation with the standard training process (Adam optimizer and a learning rate of $1e-4$, trained until convergence).
\begin{table}[b]
\caption{Comparison of our proposed curriculum learning strategies with the baselines. 
Baseline and Baseline* rows present the results of training the classifier network without and with data augmentation, respectively.
The results presented in this table are with the best-found hyperparameters regarding the portion of SSTN-processed data in the Mixed-batch approach and the number of transformation time-steps in Incremental Difficulty training. For ablation, please refer to \autoref{tab:timestep} and \autoref{tab:mixedbatch}.
}
\label{tab:main}
\centering
\begin{tabular}{lcc}
     \textbf{Method                } & \textbf{Cluttered MNIST}($\%$) & \textbf{Cluttered Fashion-MNIST}($\%$)  \\ \hline
     Baseline & 87.9 & 72.6 \\
     Baseline* & 91.2 & 83.6 \\
     Mixed-batch & 89.9  & 83.5 \\ 
     Dynamic Mixed-batch  & 94.7 & 83.7\\
     Incremental Difficulty & \textbf{95.0} & \textbf{84.9}\\
\end{tabular}
\end{table}
For the Mixed-batch experiment, we tried different setups including varying portions of SSTN-processed data in each mini-batch. The best-found result is recorded in \autoref{tab:main} and the ablation on this hyperparameter is presented in \autoref{tab:mixedbatch}.
In Dynamic Mixed-batch, initially, each mini-batch is composed of the images transformed by SSTN (easier).
Then, we gradually reduce the portion of easy-to-original by a rate of one sample per epoch.
In the Incremental difficulty experiment, we start the training with the data processed by SSTN for a maximum of $T$ time-steps.
The number of transformation steps is decreased by one every five epochs, slowly shifting the data distribution from \textit{easy} towards the original dataset.

Although both cluttered MNIST and Fashion-MNIST inherently include affine transformations (as it is used in the generation of these datasets), we observed that applying further affine transformation during the training in the context of data augmentation is effective, especially in the case of cluttered Fashion-MNIST.
According to the results in \autoref{tab:main}, we see that the Mixed-batch variant performs merely as well as standard data augmentation; however, Dynamic Mixed-batch achieves substantially better performance compared to the Mixed-Batch.
Furthermore, we observe that the  Incremental Difficulty training considerably improves the performance compared to the baselines. 
From these results, we conclude that it is important to finalize the training of the neural networks with the data distribution which is to be used at the inference phase.
We note that the accuracy improvement is more pronounced in the case of cluttered MNIST dataset;
we believe this is because the underlying SSTN trained on cluttered MNIST does a better job in processing the input and removing the clutter from the image. This is manifested in the higher performance gain achieved when using SSTN for cluttered MNIST compared to the Fashion-MNIST. Therefore, improvements in designing the STN may transfer to our proposed curriculum learning strategy.

In \autoref{tab:sched}, we present the results for Incremental Difficulty training when experimenting with three different scheduling methods. These functions are used to determine the rate at which the number of transformation time-steps is dropped from $T$ to $0$.
In Linear scheduling, the number of transformation steps is reduced by one every five epochs. 
The transformations steps in Cosine Annealing follows $\lfloor cos(\frac{epoch.\pi}{2T})\rfloor$ and Exponential Decay schedulers follows $\lfloor exp(-\frac{epoch.T}{\tau}) \rfloor$, where the hyperparameter $\tau$ determines the decay pace and is set to $30$.
We obtained the best results using simple Linear scheduling.

\begin{table}[ht]
\caption{The impact of three different scheduling functions on the Incremental Difficulty training. 
The number of transformation steps $T$ is set to $20$ in this set of experiments. 
}
\label{tab:sched}
\centering
\begin{tabular}{lcc}
     \textbf{Method} & \textbf{Cluttered MNIST}($\%$) & \textbf{Cluttered Fashion-MNIST}($\%$)  \\ \hline
     Linear Decay & \textbf{94.8} & \textbf{84.9} \\
     Cosine Annealing & 93.7 & 83.1 \\
     Exponential Decay & 94.7 & 84.7 \\
\end{tabular}
\end{table}

\subsection{Ablation}
In this part, we provide an ablation on the role of the number of transformation steps $T$ in the Incremental Difficulty as well as the impact of the portion of SSTN-processed data in Mixed-batch curriculum learning.
In \autoref{tab:timestep}, we present the results for the Incremental Difficulty method with various $T$ values (maximum number of transformation steps). 
We see that the model performance converges with about $20$ transformation steps.
\begin{table}[htb]
\caption{An ablation on impact for $T$, the maximum number of transformation steps in our Incremental Difficulty approach.}
\label{tab:timestep}
\centering
\begin{tabular}{lcc}
     \textbf{Time-steps    } & \textbf{Cluttered MNIST}($\%$) & \textbf{Cluttered Fashion-MNIST}($\%$)  \\ \hline
     T : 1 & 92.6 & 83.1\\ 
     T : 10 & 94.0& 84.4\\
     T : 20 & 94.8 & \textbf{84.9}\\
     T : 40 & \textbf{95.0} & \textbf{84.9}\\
     
\end{tabular}
\end{table}

In \autoref{tab:mixedbatch}, we experiment with different ratios of easy-to-original data in the Mixed-batch training approach.
In this approach, we obtained the best results by including $4$ and $16$ SSTN-processed samples in a mini-batch of $64$ images in cluttered MNIST and Fashion-MNIST datasets respectively.
\begin{table}[htb]
\caption{Ablation on the Mixed-batch experiment, when using a different ratio of SSTN-processed data in a mini-batch of size $64$.}
\label{tab:mixedbatch}
\centering
\begin{tabular}{lcc}
     \textbf{Ratio} & \textbf{Cluttered MNIST}($\%$) & \textbf{Cluttered Fashion-MNIST}($\%$)  \\ \hline
    
     4/64 & \textbf{89.9} & 82.9 \\
     8/84 & 89.3 & 83.2\\
     16/64 & 88.6 & \textbf{83.5}\\
     32/64 & 86.0 & 82.9\\ 

\end{tabular}
\end{table}

\section{Conclusion and Future Work}
In this paper, we presented a new approach for curriculum learning by employing SSTN~\cite{azimi2019reinforcement} to generate the easy-to-hard task ordering which is known to be advantageous to the training of neural networks.
STNs have been successful in de-cluttering the input image and modifying the input image such that it is easier for the network to classify, i.e. the classifier achieves higher accuracy~\cite{jaderberg2015spatial,azimi2019reinforcement}. 
In this work, we used this characteristic to develop a curriculum learning strategy without the need to manually define an a priori ordering of training data, by allowing the SSTN to modify all or part of the training data according to a predefined strategy. 
We applied these methods to the problem of training a classifier for cluttered versions of the MNIST and Fashion-MNIST datasets and showed that they result in increased classifier accuracy for a given architecture.
We observed that the gained improvement was more significant in the case of the cluttered MNIST, potentially due to having a more effective SSTN policy for this dataset.
Therefore, a future research direction that can directly benefit our work would be improvements in the design of the SSTN algorithm.
Furthermore, we would like to investigate how to scale the suggested method to more complex datasets and problems in the future.

\section*{Acknowledgment}
This work was supported by the TU Kaiserslautern CS PhD scholarship program, the BMBF project ExplAINN (01IS19074), and the NVIDIA AI Lab (NVAIL) program.
Further, we thank all members of the Deep Learning Competence Center at the DFKI for their feedback and support.
\bibliographystyle{splncs04}
\bibliography{egbib}
\end{document}